\documentclass[sigconf]{acmart}

\usepackage{multirow}
\usepackage{enumitem}
\usepackage{amsthm}
\usepackage{subfigure}

\theoremstyle{definition}
\newtheorem{definition}{Definition}[]

\AtBeginDocument{%
  }

\begin{document}

%% \title[short title]{full title}
\title{
    Context-Enhanced Multi-View Trajectory Representation Learning: Bridging the Gap through Self-Supervised Models
}


\author{Tangwen Qian}
\email{qiantangwen@ict.ac.cn}
\author{Junhe Li}
\email{sljhhy@gmail.com}
\affiliation{
    \institution{Institute of Computing Technology, Chinese Academy of Sciences}
    \institution{University of Chinese Academy of Sciences}
    \country{Beijing, China}
}

\author{Yile Chen}
\email{yile001@e.ntu.edu.sg}
\authornote{Corresponding Authors.}
\author{Gao Cong}
\email{gaocong@ntu.edu.sg}
\affiliation{
    % \institution{College of Computing and Data Science}
    \institution{Nanyang Technological University}
    \country{Singapore}
}

\author{Tao Sun}
\authornotemark[1]
\author{Fei Wang}
\author{Yongjun Xu}
\email{{suntao,wangfei,xyj}@ict.ac.cn}
\affiliation{
    \institution{Institute of Computing Technology, Chinese Academy of Sciences}
    \institution{University of Chinese Academy of Sciences}
    \country{Beijing, China}
}

\renewcommand{\shortauthors}{Anon et al.}
\renewcommand\footnotetextcopyrightpermission[1]{}

\begin{abstract}
    Modeling trajectory data with generic-purpose dense representations has become a prevalent paradigm for various downstream applications, such as trajectory classification, travel time estimation and similarity computation. However, existing methods typically rely on trajectories from a single spatial view, limiting their ability to capture the rich contextual information that is crucial for gaining deeper insights into movement patterns across different geospatial contexts. To this end, we propose MVTraj, a novel multi-view modeling method for trajectory representation learning. MVTraj integrates diverse contextual knowledge, from GPS to road network and points-of-interest to provide a more comprehensive understanding of trajectory data. To align the learning process across multiple views, we utilize GPS trajectories as a bridge and employ self-supervised pretext tasks to capture and distinguish movement patterns across different spatial views. Following this, we treat trajectories from different views as distinct modalities and apply a hierarchical cross-modal interaction module to fuse the representations, thereby enriching the knowledge derived from multiple sources. Extensive experiments on real-world datasets demonstrate that MVTraj significantly outperforms existing baselines in tasks associated with various spatial views, validating its effectiveness and practical utility in spatio-temporal modeling.

\end{abstract}



%   \includegraphics[width=\textwidth]{figures/method_architecture}


\maketitle

\section{Introduction}

    With the rapid advancement of web-based mobile and ubiquitous computing technologies, the acquisition of trajectory data has become increasingly prevalent, creating significant opportunities for analytical and decision-making processes in urban spaces~\cite{urbanPlanning24www1,urbanPlanning24www2,emergencyManagement24www1,intelligentLogistics24www1}. 
    To harness the potential of the growing data availability, the research of trajectory representation learning has gained considerable attention in recent years. Trajectory representation learning aims to derive effective vector representations for trajectories that can be applied in various downstream tasks, such as travel time estimation~\cite{Toast21,MMTEC2023,JGRM24}, trajectory classification~\cite{TALE21,TrajCL,START23}, and similarity computation~\cite{t2vec18,traj2simvec20,similarity24vldb}. The learned representations offer flexibility and transferability by minimizing the need for task-specific model design, while maintaining competitive performance.
    
    Although substantial progress has been made for trajectory representation learning, most existing methods focus on deriving representations within a specific spatial view. For example, previous research has proposed various methods for learning trajectory representations based on the view of GPS points~\cite{TrajFormer22,GPSmodel08}, route within road networks~\cite{PIM21,Toast21}, and check-in sequences from POIs~\cite{HMRM20,TALE21}. However, these methods leverage limited information from a single perspective, failing to capture the broader insights that can be reflected in other spatial views.

     In light of this, it is promising to integrate the complementary knowledge hidden in multiple spatial views for trajectory representation learning. Specifically, we aim to associate trajectories with three types of context information which manifests different semantic perspectives: from detailed GPS points to aligned road segments and POIs. They collectively provide insights from three key aspects, namely high resolution positions, network constraints, and region functionalities. Together, these spatial views provide a comprehensive and multi-dimensional understanding of movement behaviors, and thus enhance the development of more effective representations across views. For example, in the task of travel time estimation, the road network captures underlying structured topological information, while POIs reflect latent region attributes (e.g., commercial area), thus contributing to improved prediction accuracy. However, this integration poses several critical challenges. 
    
    \begin{figure}[t]
      \centering
      \includegraphics[width=\linewidth, height=0.37\linewidth]{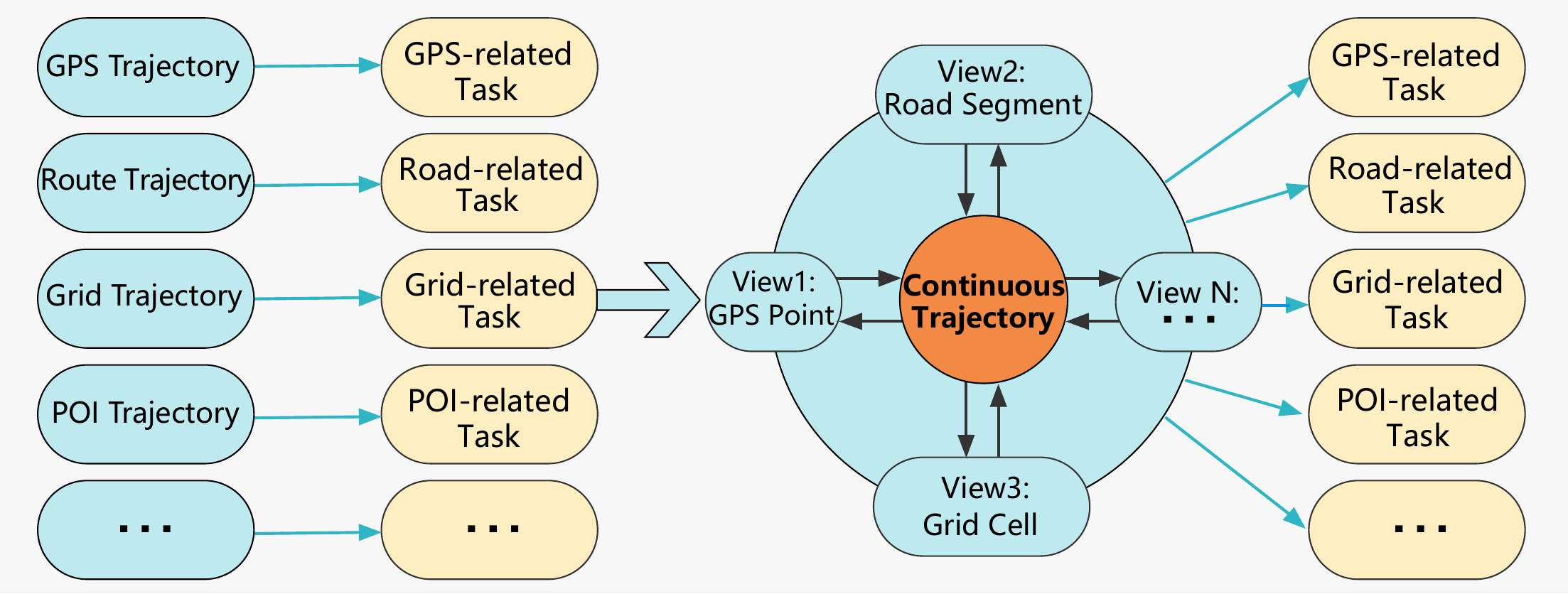}
      \caption{Motivation of multi-view trajectory representation learning. Given that GPS trajectories provide the highest sampling frequency and spatial resolution, they serve as a bridge to exhibit trajectories with various contexts, which facilitates representation learning across multiple views.}
      \label{fig:motivation}
    \end{figure}

    The first challenge is \textit{multi-view alignment}. Trajectories in different spatial views exhibit variations in their structure and format. In particular, trajectories in GPS form provide fine-grained spatial coordinates with detailed kinematic attributes but lack semantic and topological information. In contrast, trajectories aligned with road segments capture topological constraints and transition patterns, while trajectories associated with POIs offer rich semantic context related to specific areas. The inherent discrepancies across these spatial views necessitate a specific approach to align trajectories from multiple spatial views, ensuring that trajectories reflecting the same underlying movements are accurately encoded while preserving the unique spatial semantics of each view in latent space.  
    The second challenge is \textit{knowledge fusion}. Once trajectories from each view are encoded, it is critical to effectively fuse the information from different views to produce unified representations. This process is required to exchange the information across views, enabling the model to integrate complementary knowledge from each perspective and derive final representations. 

    To address the two challenges of developing a multi-view framework, we propose MVTraj, a novel self-supervised context-enhanced model for \textbf{M}ulti-\textbf{V}iew \textbf{Traj}ectory representation learning. 
    To tackle the multi-view alignment challenge, we utilize GPS trajectories as a bridge to connect diverse spatial views with self-supervised pretext tasks. Specifically, we design contrastive learning objectives to align trajectories across three views: GPS, route, and grid. This enables the model to capture and distinguish trajectory patterns across different spatial views, enhancing its capacity to learn representations with multiple spatial views. 
    To tackle the knowledge fusion challenge, we treat trajectories from different views as distinct modalities and employ a hierarchical cross-modal interaction module. This module involves utilizing cross-modal attention to capture interactions between these modalities, with twelve streams of cross-modal attention integrated through a shared Transformer to model the global context. 
    By connecting various spatial views in trajectory modeling, MVTraj generates representations that incorporate contextual information from multiple perspectives. As a result, MVTraj is not only capable of supporting trajectory-based tasks but can also accommodate downstream tasks related to route and grid representations.

    The contributions are summarized as follows: 
    \vspace{-2mm}
    \begin{itemize}
        \item We propose MVTraj, a novel method that leverages multiple spatial views to enhance trajectory representation learning. 
        To the best of our knowledge, we are the first to integrate three distinct spatial views as contextual knowledge and derive representations.
        
        \item MVTraj leverages GPS trajectories as a bridge to model these spatial views with generative and contrastive pretext tasks, addressing the multi-view alignment and knowledge fusion challenges.
        
        \item Extensive experiments on two real-world datasets demonstrate that MVTraj consistently outperforms existing baselines by a substantial margin in tasks regarding the representations from these spatial views. 
    \end{itemize}

\section{Related Work}

    In this section, we first review existing research on trajectory representation learning, followed by a discussion of studies on self-supervised learning approaches.

\subsection{Trajectory Representation Learning}

    Modeling trajectories and deriving their representations have become fundamental paradigms in the context of deep learning. 
    One direction is to derive effective trajectory representations for specific downstream tasks using supervised signals, such as trajectory similarity computation~\cite{t2vec18,traj2simvec20,t3s21}, travel time estimation~\cite{ETA20kdd,ETA20vldb,ETA20sigmod}, or route recovery~\cite{deepst20,mtrajrec21,rntraj23}. 
    While these methods demonstrate competitive performance within their respective domains, they suffer from limited generalization capabilities and often exhibit degraded performance when applied to other trajectory-specific tasks~\cite{MMTEC2023}. 

    To this end, recent methods have focused on developing generic-purpose trajectory representations that are transferrable across various downstream tasks. These methods can be broadly categorized based on the spatial views of the trajectory data they model.
    Some approaches model directly on raw GPS trajectories, such as TrajFormer~\cite{TrajFormer22}, which captures fine-grained movement patterns using raw GPS data to provide detailed kinematic attributes. 
    However, raw GPS data often contains redundancy and noise, and urban trajectories are typically constrained by the road network. Consequently, another line of research employs road network-based methods, converting GPS trajectories into route trajectories via map-matching to better reflect the underlying network structure and transition patterns~\cite{Toast21,START23,PIM21,JCRLNT22}. 
    Additionally, some methods propose to model POI trajectories, utilizing the rich semantic information associated with POIs to provide rich contextual knowledge than raw GPS data~\cite{HMRM20,TALE21}. Despite these advancements, few studies have tackled the problem of multi-view joint modeling for trajectories. One recent attempt is JGRM~\cite{JGRM24}, which integrates GPS and route trajectories to capture road network constraints and modality interactions, but it does not fully exploit the information available from POIs. 
    In contrast, our method, MVTraj, introduces a novel multi-view framework that integrates not only GPS and route views but also POIs to enrich trajectory representations. MVTraj addresses challenges such as data misalignment and structural differences by using GPS trajectories as a bridge to synchronize diverse data views. This method provides a more comprehensive understanding of spatio-temporal dynamics in trajectory data.
    
\subsection{Self-supervised Learning}

    In recent years, self-supervised learning (SSL) has been widely adopted in geospatial domain, facilitating the development of representations for POIs~\cite{POI22kdd,POI19iclr,POI23sigir}, urban regions~\cite{region17cikm,region19cikm,region22gis}, and trajectories~\cite{MMTEC2023,ssl-www1,ssl-www2}. The interest of applying SSL techniques to geospatial data is driven by their intrinsic advantages to derive effective representations for urban analytical units (e.g., trajectories) without the need for extensive labels~\cite{spatial_ssl}. For trajectory data discussed in the paper,
    SSL is typically divided into two main categories of pretext tasks: generative and contrastive tasks~\cite{ssl-survey22,ssl-survey23,ssl-survey24}. 

    Generative pretext tasks in trajectory learning often involve masked-token prediction or next-token prediction~\cite{t2vec18,Trembr20,pt2vec23,ssl-generative}. 
    Inspired by the success of masked language models in natural language processing, these tasks have been adapted for geospatial data by masking sequences of road segments, leveraging the connected structure of road networks to provide richer contextual information for representation learning~\cite{Toast21,JGRM24}. 
    Next-token prediction, another generative strategy, trains models to predict subsequent points or segments in a trajectory~\cite{START23}, which combines this approach with other pretext tasks to enhance trajectory understanding. 

    Contrastive pretext tasks, on the other hand, focus on constructing positive and negative samples through data augmentation, allowing models to learn by contrasting different views of the same data~\cite{tpr22,PIM21,JCRLNT22,ssl-constrastive}. 
    For instance, LightPath~\cite{LightPath23} downsamples trajectories and uses sparse encoders to create distinct views for contrastive learning. 
    MMTEC~\cite{MMTEC2023} introduces a multi-view trajectory framework that integrates both semantic and continuous spatio-temporal information into embeddings, utilizing inductive attention and neural differential equations to generate discrete and continuous data views. 
    JGRM~\cite{JGRM24} performs contrastive learning across different perspectives, i.e., GPS and route trajectory, ensuring robust cross-view learning.

    Unlike prior studies, MVTraj further integrates POIs with another spatial view alongside GPS and route information. 
    This integration is achieved through a novel use of GPS trajectories as a bridge for aligning diverse data types, enabling more effective multi-view learning for trajectories. 

\section{Problem Formulation}

    In this section, we introduce the fundamental definitions and the problem formulation in this paper. 

    \begin{definition}(\textbf{Trajectory}).
        A trajectory $\mathcal{T}$ of length $\vert \mathcal{T} \vert$ is defined as a sequence of sampled points, denoted as $\mathcal{T} = \{ ( {pos}_i, {t}_i ) \}_{i=1}^{\vert \mathcal{T} \vert}$, where ${pos}_i$ and ${t}_i$ represent the geographical position and timestamp at the $i$-th sample point in the sequence. 
    \end{definition}
    
    Built on this basic form, GPS trajectories can be enriched with various context information, serving as a bridge to multiple spatial views, each providing a distinct semantic perspective~\cite{scaleDefinition23}. 
    In this work, we consider three specific views representing different data modalities: 
    (1) The GPS point view $p$, which offers the highest spatial resolution; 
    (2) The road network view $r$, which aligns movements with road segments within a network structure; 
    (3) The grid view $g$, which aggregates movements into coarser, grid-based urban areas. 
    
    \begin{definition}(\textbf{GPS Trajectory}).
        A GPS trajectory $\mathcal{T}^p$ is a sequence of sampled GPS points, denoted as: 
        \begin{displaymath}
            \mathcal{T}^{p} = \{ ( {lat}_i, {lon}_i, {t}_i ) \}_{i=1}^{\vert \mathcal{T}^{p} \vert}
        \end{displaymath}
        where $lat_i$ and $lon_i$ refer to the latitude and longitude at the $i$-th point, and $t_i$ is the associated timestamp. 
    \end{definition}

    \begin{definition}(\textbf{Route Trajectory}).
        A route trajectory $\mathcal{T}^r$ is a sequence of adjacent road segments, denoted as: 
        \begin{displaymath}
            \mathcal{T}^{r} = \{ ( {v}_i, {t}_i ) \}_{i=1}^{\vert \mathcal{T}^{r} \vert}
        \end{displaymath}
        where $v_i \in \mathcal{V}$ is the $i$-th road segment in the trajectory, and ${t}_i$ represents its corresponding arrival timestamp. $\mathcal{V}$ is a set of vertices representing road segments in the road network $G=( \mathcal{V}, \mathcal{E})$, and $\mathcal{E} \subseteq \mathcal{V} \times \mathcal{V}$ is a set of edges representing intersections between road segments, 
    \end{definition}

    To obtain a route trajectory $\mathcal{T}^r$, the GPS trajectory $\mathcal{T}^p$ is aligned with the road network $G$ through the map-matching algorithm~\cite{fastMapMatch18}. 

    \begin{definition}(\textbf{Grid Trajectory}).
        A grid trajectory $\mathcal{T}^g$ is a sequence of traversed grid cells, denoted as:
        \begin{displaymath}
            \mathcal{T}^{g} = \{ ( {grid}_i, {sem}_i, {t}_i ) \}_{i=1}^{\vert \mathcal{T}^{g} \vert}
        \end{displaymath}
        where ${grid}_i$ is the $i$-th grid cell in the trajectory, ${sem}_i$ represents the semantic representation associated with the grid cell, and $t_i$ is the arrival timestamp of that cell. 
    \end{definition}

    The grid trajectory $\mathcal{T}^g$ is derived from the GPS trajectory $\mathcal{T}^p$ by partitioning the geographic space of interest into uniform, non-overlapping grid cells. Since points-of-interest (POI) naturally exhibit semantic characteristics and reflect the functional role of a region, we propose utilizing POIs to derive the semantic representation $sem_i$ for each grid.   

    These three spatial views capture distinct and complementary aspects of the contextual meaning in traversed locations, not only for trajectories but also for the data modality specific to each view. The GPS trajectory $\mathcal{T}^p$ provides position information with the highest spatial resolution but lacks inherent semantic context.
    The route trajectory $\mathcal{T}^r$ emphasizes the structural and transition patterns of movement within road segments. The grid trajectory $\mathcal{T}^{g}$ offers an enriched understanding by incorporating POIs to reflect the functional roles of different areas. Together, these views, with the trajectory as the integrating factor, provide a comprehensive and multi-dimensional context that supports better representation learning for the data modality in each spatial view.

 \noindent\textbf{Problem Statement}.
    Given a trajectory dataset $\mathcal{D} = \{ \mathcal{T}_i \}_{i=1}^{\vert \mathcal{D} \vert}$, our target is to learn generic-purpose representations for trajectories, road segments and grid cells that can be effectively utilized in a variety of downstream tasks across these views. The learned representations require minimal task-specific adaptations, ensuring flexibility and broad applicability across tasks. 

\section{Methodology}

    In this section, we introduce the overview of MVTraj and describe the details of its components. 

    \begin{figure*}[ht]
      \centering
      \includegraphics[width=0.9\linewidth, height=0.4\linewidth]{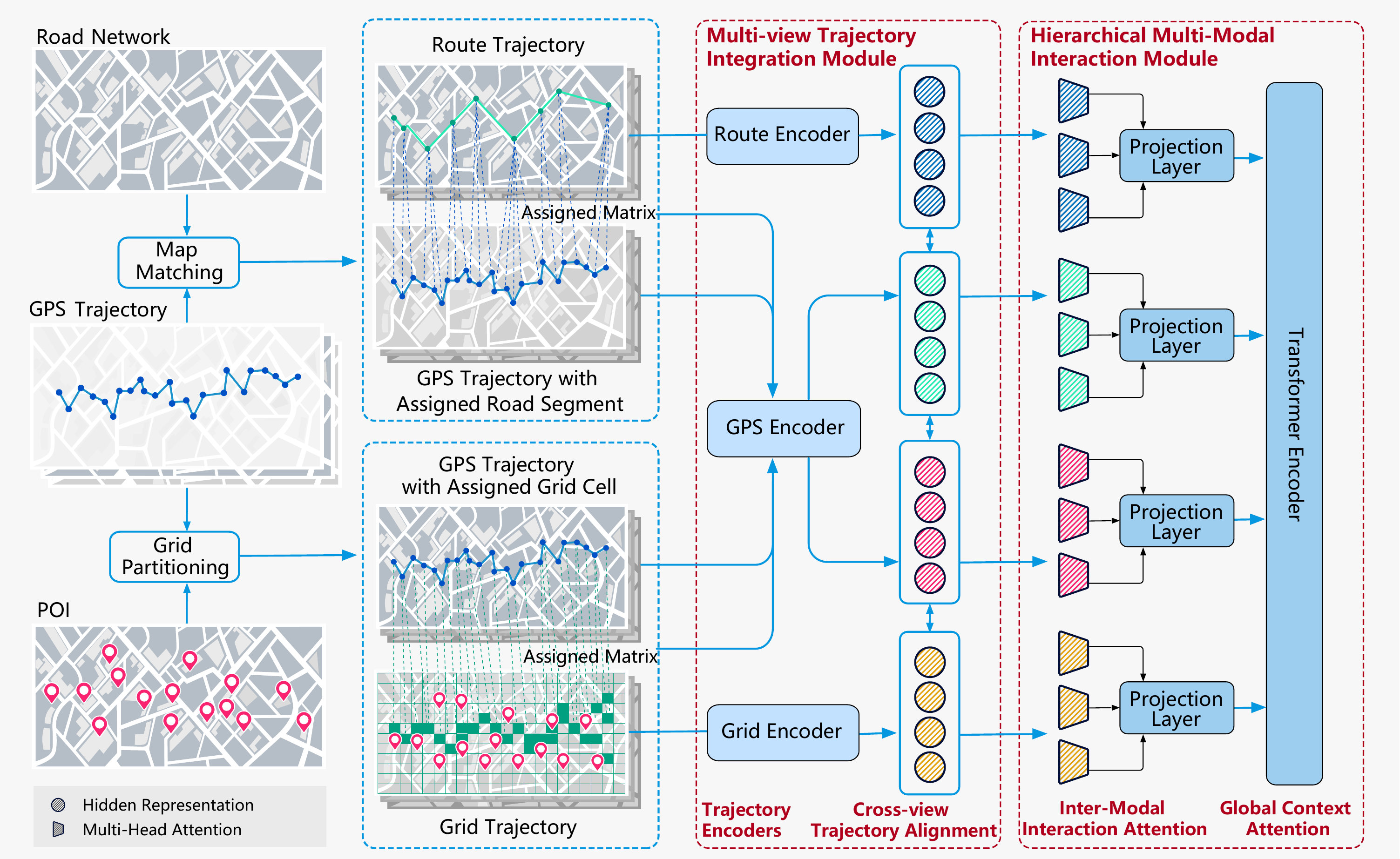}
      \caption{An overview of MVTraj. MVTraj consists of two key modules: a multi-view trajectory integration module, and a hierarchical multi-modal interaction module.}
      \label{fig:model_architecture}
    \end{figure*}

\subsection{Overview}
    
    The framework of MVTraj, illustrated in Fig.~\ref{fig:model_architecture}, is designed to treat trajectories as data-centric entities, and leverage their contextual information embedded in multiple spatial views to derive effective representations that can be applied to tasks associated with each respective spatial view. 
    
    MVTraj consists of two key modules: (1) a multi-view trajectory integration module to align different spatial views, and (2) a hierarchical multi-modal interaction module for effective knowledge fusion across these views. Specifically, for multi-view alignment, we treat GPS trajectories as a bridge to connect diverse spatial views, and apply contrastive learning objectives to align trajectories from three views: GPS, route, and grid. This alignment process allows the model to capture informative movement patterns from each view while maintaining a unified representation space. For knowledge fusion, we consider trajectories from different views as distinct modalities and employ a hierarchical multi-modal interaction mechanism, using inter-modal interaction attention and global context attention to capture interactions between these modalities. This module produces the representations that integrate complementary trajectory knowledge from these spatial views.

\subsection{Multi-view Trajectory Integration Module}

    This module consists of three trajectory encoders, each dedicated to a different spatial view, and a cross-view trajectory alignment component. The encoders are responsible for generating trajectory representations from three different spatial views: route, grid, and GPS. Once these representations are obtained, the cross-view alignment component ensures that the informative patterns from each spatial view are captured in their respective representations. At the same time, it aligns these representations in a shared latent space while preserving distinct characteristics associated with each spatial view.

\subsubsection{Trajectory Encoders}

    There are three specialized encoders for each spatial view, i.e., route encoder, GPS encoder, and grid encoder. 
    Each encoder is designed to capture unique features related to its respective spatial view, including its spatial properties (e.g., road networks, GPS points, and grid cells) and temporal dynamics (e.g., travel times and periodicity). 

    For the route view, the route encoder processes trajectories constrained by the road network and dynamic traffic conditions.
    To capture the spatial structure of the road network, the graph attention network~\cite{GAT} is employed for updating road segment embeddings based on observed trajectories traversing the network. The embedding update for each road segment $v \in \mathcal{V}$ is represented by:  
    \begin{equation}
        \boldsymbol{z}_{v} = \sigma(\sum_{u \in N(v)} \alpha_{vu} \boldsymbol{g}_{u} \boldsymbol{W})
    \end{equation}
    where $N(v)$ denotes the neighboring road segments of $v$, $\alpha_{vu}$ is the learned attention coefficient between road segments $v$ and $u$, $\boldsymbol{g}_{u}$ denotes the initial embedding of road segment $u$, and $\boldsymbol{W}$ is a learnable weight matrix. The non-linearity is introduced by the activation function $\sigma$ allowing the model to capture complex relationships between connected road segments. 
    The temporal aspect of each road segment is characterized by both discrete features (e.g., minute of the day, day of the week) and continuous features (e.g., actual travel time). The final temporal representation $\boldsymbol{t}_v$ for each road segment is computed by summing these discrete and continuous feature components. Specifically, the temporal embedding is added to the spatial embedding to form a complete road segment representation $\boldsymbol{r}_{v} = \boldsymbol{z}_v + \boldsymbol{t}_v$. 
    To model the dependencies between road segments along a given route, we utilize a transformer-based architecture. The segment-level representations and the trajectory-level representation are obtained as follows: 
    \begin{equation}
        \boldsymbol{h}_{\mathcal{T}^r}, \boldsymbol{h}_{\mathcal{S}^r} = RouteEncoder(\mathcal{T}^r, \boldsymbol{r}_v)
    \end{equation}
    where $\boldsymbol{h}_{\mathcal{S}^r}$ represents the segment-level (i.e., token) representations of the road segments within the route, and $\boldsymbol{h}_{\mathcal{T}^r}$ represents the aggregated trajectory-level representation obtained through a mean pooling operation over the segment-level representations.

    For the GPS view, the GPS encoder processes trajectories with GPS points in a hierarchical structure. Specifically, GPS trajectories are divided into sub-trajectories where points correspond to either the same road segment $\mathcal{T}^{p|r}$ or grid cell $\mathcal{T}^{p|g}$. This correspondence is described via a binary assignment matrix $B_{p}^{r} \in \mathbb{R}^{|\mathcal{T}^{p}|\times |\mathcal{V}|}$, where a value of 1 indicates that a GPS point is located on the corresponding road segment. Similarly, the assignment matrix $B_{p}^{g}$ is produced to describe the correspondence between GPS points and grid cells. Subsequently, the points within each sub-trajectories, as well as the sub-trajectories themselves, are encoded using a bidirectional GRU respectively, forming a two-level model architecture:
    \begin{equation}
    \begin{aligned}
        \boldsymbol{h}_{\mathcal{T}^{p|r}}, \boldsymbol{h}_{\mathcal{S}^{p|r}} = GPSEncoder(\mathcal{T}^{p|r}, B_{p}^{r}) \\
        \boldsymbol{h}_{\mathcal{T}^{p|g}}, \boldsymbol{h}_{\mathcal{S}^{p|g}} = GPSEncoder(\mathcal{T}^{p|g}, B_{p}^{g}) 
    \end{aligned}
    \end{equation}
    where $\boldsymbol{h}_{\mathcal{T}^{p|r}}$ and $\boldsymbol{h}_{\mathcal{T}^{p|g}}$ represent GPS trajectory representations aligned with road segments and grid cells, respectively, while $\boldsymbol{h}_{\mathcal{S}^{p|r}}$ and $\boldsymbol{h}_{\mathcal{S}^{p|g}}$ denote the corresponding segment-level representations for the road segments and grid cells. 
    
    For the grid view, the grid encoder captures the spatial relationships between grid cells and integrates semantic features from POIs to reflect the functional characteristics of the area. 
    To model the dependencies between grid cells along a given grid trajectory, we utilize a transformer-based architecture:
    \begin{equation}
    \begin{aligned}
        \boldsymbol{h}_{\mathcal{T}^{g}}, \boldsymbol{h}_{\mathcal{S}^{g}} = GridEncoder(\mathcal{T}^{g}, Sem(\mathcal{T}^{g}))
    \end{aligned}
    \end{equation}
    where $\boldsymbol{h}_{\mathcal{T}^{g}}$ represents the grid trajectory representation, $\boldsymbol{h}_{\mathcal{S}^{g}}$ denotes the segment-level representation corresponding to the grid trajectory, 
    and $Sem(\mathcal{T}^{g})$ denotes the semantic representation for the grid trajectory, calculated as the weighted sum of all POI categories present within it based on their frequencies, where each POI category is associated with a learnable embedding vector that captures its characteristics.

\subsubsection{Cross-view Trajectory Alignment}

    Building on the trajectory encoders, which generate trajectory representations that capture the unique patterns in each spatial view, 
    we further ensure these representations are consistent in the latent space across different spatial views for the subsequent fusion of their distinct characteristics in each view. To achieve this, we adopt techniques from cross-modal retrieval studies~\cite{contrastiveLearning}, which applies contrastive learning to align trajectory representations across spatial view pairs: (1) route trajectories ($\mathcal{T}^{r}$) and GPS trajectories assigned to road segments ($\mathcal{T}^{p|r}$). (2) grid trajectories ($\mathcal{T}^{g}$) and GPS trajectories assigned to grid cells ($\mathcal{T}^{p|g}$). (3) GPS trajectories assigned to road segments ($\mathcal{T}^{p|r}$) and assigned to grid cells ($\mathcal{T}^{p|g}$). 

    We adopt contrastive learning to distinguish the positive trajectory pairs that represent the same underlying trajectories across different views, from negative pairs, which are randomly sampled. The loss for a trajectory $\mathcal{T}^a$ between its representations from view $i$ and $j$ is expressed as: 
    \begin{equation}
        \mathcal{L}_{pair}^{i,j}(\mathcal{T}^a) = - log \frac{exp(sim(\boldsymbol{h}_{\mathcal{T}^a}^{i}, \boldsymbol{h}_{\mathcal{T}^a}^{j})/\tau)}{\sum_{\mathcal{T}^b \in \mathcal{B}} exp(sim(\boldsymbol{h}_{\mathcal{T}^a}^{i}, \boldsymbol{h}_{\mathcal{T}^b}^{j})/\tau) }
    \end{equation}
    where $\mathcal{T}^b$ represents other trajectories in the batch $\mathcal{B}$, $sim(\cdot)$ is the cosine similarity between two representations, and $\tau$ is a temperature hyperparameter. 

    Then total multi-view alignment loss for a trajectory $\mathcal{T}^a$ is the aggregation of the three trajectory view pairs: 
    \begin{equation}
        \mathcal{L}_{align} (\mathcal{T}^a) = \mathcal{L}_{pair}^{r, p|r}(\mathcal{T}^a) + \mathcal{L}_{pair}^{g, p|g}(\mathcal{T}^a) + \mathcal{L}_{pair}^{p|r, p|g}(\mathcal{T}^a)
    \end{equation}
    Finally, the overall loss for the entire batch $\mathcal{B}$ is obtained by averaging the loss over all trajectories in the batch:
    \begin{equation}
        \mathcal{L}_{align}^{multi} = \frac{1}{\vert \mathcal{B} \vert} \sum_{\mathcal{T}^a \in \mathcal{B}} \mathcal{L}_{align} (\mathcal{T}^a)
    \end{equation}
    
\subsection{Hierarchical Multi-Modal Interaction Module}

    After obtaining the representations from the multi-view trajectory integration module, the hierarchical multi-modal interaction module is designed to effectively fuse knowledge across multiple views. This module treats the four trajectory representations, namely route trajectory $\boldsymbol{h}_{\mathcal{T}^r}$, GPS trajectory assigned to road segments $\boldsymbol{h}_{\mathcal{T}^{p|r}}$, grid trajectory $\boldsymbol{h}_{\mathcal{T}^g}$, and GPS trajectory assigned to grid cells $\boldsymbol{h}_{\mathcal{T}^{p|g}}$ as distinct modalities, thereby extracting multi-dimensional information from these views. 

\subsubsection{Inter-Modal Interaction Attention}

    The four types of trajectory representations, derived from different spatial perspectives, are first processed through an inter-modal interaction attention mechanism, which establishes twelve dedicated attention streams to capture pairwise interactions between these modalities. 

    For each modality, three independent attention operations are performed, using the remaining three modalities as keys and values.
    Specifically, the multi-head attention computation for a given query modality $\mathcal{T}^a$ with respect to a key-value modality $\mathcal{T}^b$ can be expressed as: 
    \begin{equation}
        \boldsymbol{O}_a^b = MultiHead(\mathcal{T}^a, \mathcal{T}^b, \mathcal{T}^b) = softmax(\frac{Q_a K_b^{\top}}{\sqrt{d_k}} ) V_b
    \end{equation}
    where $Q_a = W_q [\boldsymbol{h}_{\mathcal{T}^a}, \boldsymbol{h}_{\mathcal{S}^a}]$ is the projected query from modality $\mathcal{T}^a$, $K_b = W_k [\boldsymbol{h}_{\mathcal{T}^b}, \boldsymbol{h}_{\mathcal{S}^b}]$ and $V_b = W_v [\boldsymbol{h}_{\mathcal{T}^b}, \boldsymbol{h}_{\mathcal{S}^b}]$ are the projected value from modality $\mathcal{T}^b$. For each query modality $\mathcal{T}^a$ corresponding to a particular spatial view, this mechanism is performed three times, once for each of the other modalities $\mathcal{T}^b \in \{ \mathcal{T}^r, \mathcal{T}^{p|r}, \mathcal{T}^g, \mathcal{T}^{p|g} \} \setminus \mathcal{T}^a$, enabling pairwise interaction between two spatial views. 
    
    The outputs from the three attention streams are aggregated to form the enhanced representation for each modality 
    $\boldsymbol{O}_a = \sum_{\mathcal{T}^b} \boldsymbol{O}_a^b$.
    By leveraging this process, the model effectively integrates knowledge from different spatial views, enriching each modality’s representation with complementary insights from the others. This enables the model to achieve a more comprehensive understanding of movement behaviors and spatial-temporal relationships from multiple aspects. 

\subsubsection{Global Context Attention}

    Following the inter-modal interaction attention module, the enriched representations from all modalities are further consolidated using a global context attention mechanism. In this phase, a shared Transformer is employed to model the global context and interactions across all trajectory modalities. 

    The shared Transformer processes concatenated outputs from inter-modal interaction attention streams, refining fused trajectory representations with global information from all spatial views: 
    \begin{equation}
        [\boldsymbol{E}_{r}, \boldsymbol{E}_{p|r}, \boldsymbol{E}_{g}, \boldsymbol{E}_{p|g}] = TransEncoder([\boldsymbol{O}_{r}, \boldsymbol{O}_{p|r}, \boldsymbol{O}_{g}, \boldsymbol{O}_{p|g}])
    \end{equation}

    In addition, we employ a masked language modeling (MLM) task to train the hierarchical cross-modal interaction module. The MLM task operates by randomly masking portions of the trajectory data and requiring the model to predict the masked elements, thereby forcing the model to learn generalized representations of the trajectory views. 
    Formally, the MLM loss is defined as the negative log-likelihood of correctly predicting the masked tokens $m_i$:
    \begin{equation}
        \mathcal{L}_{MLM}^{\mathcal{T}} = - \sum_{m_i \in \boldsymbol{M}^{\mathcal{T}}} log P(\boldsymbol{e}_{m_i} | \boldsymbol{E}_{\setminus m_i})
    \end{equation}
    where $P(\boldsymbol{e}_{m_i} | \boldsymbol{E}_{\setminus m_i})$ represents the probability of model correctly predicting the masked token $\boldsymbol{e}_{m_i}$ given the unmasked context $\boldsymbol{E}_{\setminus m_i}$. 
    $\boldsymbol{M}^{\mathcal{T}} = \{ m_1, m_2, \cdots, m_{\vert \boldsymbol{M}^{\mathcal{T}} \vert} \}$ denotes the set of masked elements. 

    Given the four trajectory modalities $(\mathcal{T}^r, \mathcal{T}^{p|r}, \mathcal{T}^g, \mathcal{T}^{p|g})$, the total MLM loss across all modalities is computed as:
    \begin{equation}
        \mathcal{L}_{MLM}^{multi} = \mathcal{L}_{MLM}^{\mathcal{T}^{r}} + \mathcal{L}_{MLM}^{\mathcal{T}^{p|r}} + \mathcal{L}_{MLM}^{\mathcal{T}^{g}} + \mathcal{L}_{MLM}^{\mathcal{T}^{p|g}}
    \end{equation}

    Finally, the total loss combines the multi-view alignment loss $\mathcal{L}_{align}^{multi}$ and the MLM loss $\mathcal{L}_{MLM}^{multi}$:
    \begin{equation}
    \label{eq:total_loss}
        \mathcal{L} = w_1 \mathcal{L}_{align}^{multi} + w_2 \mathcal{L}_{MLM}^{multi}
    \end{equation}
    where $w_1$ and $w_2$ are hyperparameters that balance the two losses during training. 
    
\section{Experiments}

    To evaluate the performance of MVTraj, we conduct extensive experiments to answer the following research questions: 
    \begin{itemize}[leftmargin=*]
        \item \textbf{RQ1}: How does MVTraj perform compared to state-of-the-art trajectory representation learning models? (Sec.~\ref{subsec:exp_rq1})
        \item \textbf{RQ2}: How do self-supervised objectives affect the effectiveness of MVTraj? (Sec.~\ref{subsec:exp_rq2})
        \item \textbf{RQ3}: How does each component of MVTraj contribute to the overall performance? (Sec.~\ref{subsec:exp_rq3})
        \item \textbf{RQ4}: How efficient is MVTraj in terms of inference time? (Sec.~\ref{subsec:exp_rq4})
    \end{itemize}

\subsection{Experimental Setup}\label{subsec:exp_setup} 

\subsubsection{Datasets and Preprocessing}

    We conduct experiments on two real-world datasets, which have been widely used in previous studies on the evaluation of trajectory representations~\cite{JGRM24,LightPath23,MMTEC2023,JCRLNT22}, i.e., Chengdu and Xi'an.     
    Each dataset contains taxi trip \textbf{GPS trajectories} collected from the central areas of the respective cities. 

    To obtain road-network constrained \textbf{route trajectories}, we collect road network data from \textit{Open Street Map}\footnote{https://www.openstreetmap.org/} and perform the map matching~\cite{fastMapMatch18} procedure to align GPS points with road segments. 
    To derive \textbf{grid trajectories}, we collect POI data\footnote{http://geodata.pku.edu.cn} and normalize all POIs in the grid cell to represent the semantic information of the grid cell. 
    Due to space constraints, further details are provided in Appendix~\ref{subsec:appendix_datasets} and Appendix~\ref{subsec:appendix_implementation_details}. 

\subsubsection{Downstream Tasks and Evaluation Metrics}

    The model is evaluated on a total of four tasks across the three spatial views, including two segment-level tasks and two trajectory-level tasks. For each task, we employ a variety of evaluation metrics that measure model performance in the respective tasks. 
    Due to space limitations, the full details of evaluation metrics, including definitions and computation methods, are provided in Appendix~\ref{subsec:appendix_evaluation_metrics} for thorough understanding. 

    \textit{Segment-Level Tasks} evaluate the road segment representations derived from the route view. In these tasks, representations of the same road segments across different trajectories are averaged to create static representations, which serve as the input data for simple task-specific components (e.g., MLP)~\cite{JGRM24,START23}. 
    \begin{itemize}
        \item \textbf{Road Label Classification}: This task aims to classify road segments by treating their road types as labels. The four most frequently occurring road types (i.e., primary, secondary, tertiary, and residential) are selected to evaluate the road segment representations. The classification performance is measured using Micro-F1 and Macro-F1 scores. 

        \item \textbf{Road Speed Inference}: This task aims to infer the average speed on road segments, derived from GPS trajectory data. 
        Regression performance is measured using Mean Absolute Error (MAE) and Root Mean Squared Error (RMSE) with the ground-truth data.  

    \end{itemize}

    \textit{Trajectory-Level Tasks} evaluate the effectiveness of trajectory representations using two scenarios. Similarly, in these tasks, the trajectory representations are utilized as the input for simple task-specific components. 

        \begin{itemize}
        \item \textbf{Travel Time Estimation}: This task aims to predict the travel time for given route trajectories. The input consists of route trajectories with time information masked to prevent information leakage. The performance for the predicted travel time is evaluated using MAE and RMSE.
        
        \item \textbf{Destination Grid Prediction}: This task aims to predict the destination grid for a trajectory based on its representation derived from the grid view. The classification performance is evaluated using top-k accuracy denoted as $Acc@k$. 
        
    \end{itemize}
    
    It is important to note that for these tasks, only the classification or regression components are trained, while all other parameters remain fixed. This ensures that the evaluation reflects the quality of the learned representations, eliminating any confounding influence from further model updates.

\subsubsection{Compared Methods}

    To evaluate the effectiveness of our proposed method, following previous studies~\cite{START23,JGRM24}, we compare it against several methods, including the state-of-the-art methods for generic-purpose trajectory representation learning, under the same experimental settings mentioned in the task descriptions: 
    \begin{itemize}
        \item Random: it initializes trajectory representations randomly to serve as a baseline for comparison.
        \item Word2Vec~\cite{word2vec13}: it employs the skip-gram model to generate trajectory representations based on co-occurrence statistics. 
        \item Node2Vec~\cite{node2vec16}: it explores the neighborhood of a node through the random walk, thereby capturing both local and global structural properties of the network.
        \item Transformer~\cite{Transformer}: it leverages a self-attention mechanism within an encoder-decoder architecture to capture complex dependencies in sequential data. 
        \item BERT~\cite{BERT}: it is designed for pre-training deep bidirectional representations by conditioning jointly on both left and right contexts in all layers.
        \item START~\cite{START23}: it introduces a trajectory encoder that integrates travel semantics with temporal continuity, enhanced by two self-supervised tasks. 
        \item JGRM~\cite{JGRM24}: it combines GPS traces with route traces to model road network constraints and modalities, offering a nuanced understanding of trajectory data.
    \end{itemize}

    The detailed implementation of these methods is provided in Appendix~\ref{subsec:appendix_implementation_details} for further clarity.
    
\subsection{Performance Comparison (RQ1)}\label{subsec:exp_rq1}

    The performance of various methods is summarized in Table~\ref{tab:RQ1} on Xi'an and Chengdu datasets.

    Baseline models like JGRM perform well on various tasks, with JGRM achieving MAE 87.87 in travel time estimation, demonstrating its feature-capturing ability. However, traditional methods such as Word2Vec and Node2Vec struggle in complex tasks that require broader contextual understanding, as they focus on local or sequential information. For instance, Word2Vec performs adequately in road label classification but falls short in tasks like travel time estimation, which requires global spatial dependencies.

    In contrast, our proposed model, MVTraj, significantly outperforms all baselines across tasks. In Chengdu, MVTraj reduces MAE by 81.85\% and RMSE by 53.62\% in travel time estimation compared to JGRM. This improvement underscores the power of integrating multiple spatial views, allowing MVTraj to better capture intricate spatial patterns in trajectories.

    Furthermore, the incorporation of diverse spatial perspectives not only boosts performance on grid-based tasks, such as destination grid prediction, but also enhances MVTraj's generalization across various tasks and geographical areas. Unlike many baseline models that struggle with grid-based tasks due to their single-view focus, MVTraj's multi-dimensional approach allows it to handle a wider array of challenges effectively. This is reflected in consistent performance improvements in both datasets. 

    Overall, these findings demonstrate that MVTraj not only outperforms existing methods but also addresses the limitations of traditional approaches by employing a comprehensive strategy for trajectory representation learning.

    \begin{table*}
        \caption{Overall Performance on Segment-Level and Trajectory-Level Tasks in Xi'an and Chengdu.}
        \label{tab:RQ1}
        \begin{tabular}{c|c|cc|cc|cc|cc}
            \toprule
             & \multirow{2}*{Method} & \multicolumn{2}{|c}{Road Speed Inference} & \multicolumn{2}{|c}{Road Label Classification} & \multicolumn{2}{|c}{Travel Time Estimation} & \multicolumn{2}{|c}{Destination Grid Prediction} \\
             & & MAE $\downarrow$ & RMSE $\downarrow$ & Micro-F1 $\uparrow$ & Macro-F1 $\uparrow$ & MAE $\downarrow$ & RMSE $\downarrow$ & Acc@1 $\uparrow$ & Acc@5 $\uparrow$ \\
            \midrule
            \multirow{9}*{Xi'an} & Random   & 3.2986 & 4.2090 & 0.4680 & 0.3087 & 120.9861 & 153.4056 & X & X \\
            & Word2vec    & 3.2119 & 4.1833 & 0.6525 & 0.6267$\ddagger$ & 89.5472$\ddagger$  & 122.3465$\ddagger$ & X & X \\
            & Node2vec    & 3.3126 & 4.2404 & 0.4387 & 0.2938 & 91.5226  & 124.4122 & X & X \\
            & Transformer & 3.3114 & 4.2468 & 0.4305 & 0.3645 & 91.3093  & 124.1358 & X & X \\
            & BERT        & 3.2154 & 4.1610 & 0.6780$\ddagger$ & 0.6251 & 90.2442  & 123.2867 & X & X \\
            % GAE       & 3.2496 & 4.1794 & 0.4620 & 0.4360 & 90.2352  & 122.9764 & X & X \\
            % Toast     & 3.1145 & 4.0025 & 0.7055 & 0.6606 & 92.9093  & 129.3365 & X & X \\
            % Trember   & 3.2052 & 4.1269 & 0.6627 & 0.6212 & 98.8188  & 134.7582 & X & X \\ 
            % JCRLNT    & 3.1651 & 4.0864 & 0.6090 & 0.5179 & 100.8771 & 133.8522 & X & X \\ 
            & START       & 3.1199$\ddagger$ & 4.0072$\ddagger$ & 0.4413 & 0.3575 & 118.0605 & 162.0801 & X & X \\
            & JGRM        & 2.6878$\dagger$ & 3.6039$\dagger$ & 0.7745$\dagger$ & 0.7622$\dagger$ & 87.8708$\dagger$  & 119.9921$\dagger$ & X & X \\
            % \midrule
            & MVTraj      & \textbf{2.5181} & \textbf{3.4623} & \textbf{0.8290} & \textbf{0.8159} & \textbf{54.9044}  & \textbf{85.3847}  & \textbf{0.6630} & \textbf{0.8154} \\ 
            & Improvement & 6.74\% & 4.09\% & 7.04\% & 7.05\% & 60.04\% & 40.53\% & X & X \\
            
            \midrule
            \multirow{9}*{Chengdu} & Random   & 3.5592 & 4.6376 & 0.4363 & 0.3152 & 112.3310 & 141.6182 & X & X \\
            & Word2vec    & 3.4692 & 4.5591 & 0.5857$\ddagger$ & 0.5767$\ddagger$ & 85.4754$\ddagger$  & 113.8926$\ddagger$ & X & X \\
            & Node2vec    & 3.5408 & 4.5997 & 0.5535 & 0.5306 & 85.9276  & 114.4905 & X & X \\
            & Transformer & 3.5856 & 4.6637 & 0.3753 & 0.3460 & 88.3027  & 117.2306 & X & X \\
            & BERT        & 3.5155 & 4.6091 & 0.5516 & 0.5363 & 86.8267  & 115.4532 & X & X \\
            % GAE       & 3.2870 & 4.2134 & 0.4373 & 0.3805 & 90.2352  & 122.9764 & X & X \\
            % Toast     & 3.3201 & 4.3777 & 0.6276 & 0.6195 & 86.0053  & 114.2109 & X & X \\
            % Trember   & 3.3955 & 4.4470 & 0.6110 & 0.6059 & 90.9035  & 119.9026 & X & X \\ 
            % JCRLNT    & 3.4410 & 4.5016 & 0.5169 & 0.4660 & 100.1113 & 129.5910 & X & X \\ 
            & START       & 3.3396$\ddagger$ & 4.3490$\ddagger$ & 0.3526 & 0.1869 & 112.0348 & 148.3855 & X & X \\
            & JGRM        & 2.8257$\dagger$ & 3.8198$\dagger$ & 0.7198$\dagger$ & 0.7228$\dagger$ & 82.8468$\dagger$  & 110.3405$\dagger$ & X & X \\
            % \midrule
            & MVTraj      & \textbf{2.6972} & \textbf{3.7140} & \textbf{0.7206} & \textbf{0.7326} & \textbf{48.5581}  & \textbf{71.8248}  & \textbf{0.7927} & \textbf{0.9105} \\ 
            & Improvement & 4.76\% & 2.85\% & 1.11\% & 1.36\% & 81.85\% & 53.62\% & X & X \\
            \bottomrule
        \end{tabular}
        \par{* The metric with $\uparrow$ means that larger is better, and otherwise. \textbf{Bold} denotes the best result, $\dagger$ and $\ddagger$ denotes the second and third best result. }
    \end{table*}

\subsection{Model Analysis (RQ2)}\label{subsec:exp_rq2}

    We evaluate the impact of self-supervised objectives used for pre-training on model performance by conducting experiments on the travel time estimation task on the Xi'an dataset. 
    Specifically, we compare two strategies for model training: 
    \begin{itemize}
        \item Pre-train: This strategy corresponds to the original MVTraj, where self-supervised objectives are employed for pre-training, followed by the fine-tuning only on the regression head on the travel time estimation task. 
        \item No Pre-train: This strategy represents the variant trained in an end-to-end manner, where both the trajectory encoders and the regression head are randomly initialized and optimized jointly from scratch through the supervised labels from this task. 
    \end{itemize}

\begin{figure}
        \centering
        \subfigure[MAE]{
            \includegraphics[width=0.47\linewidth]{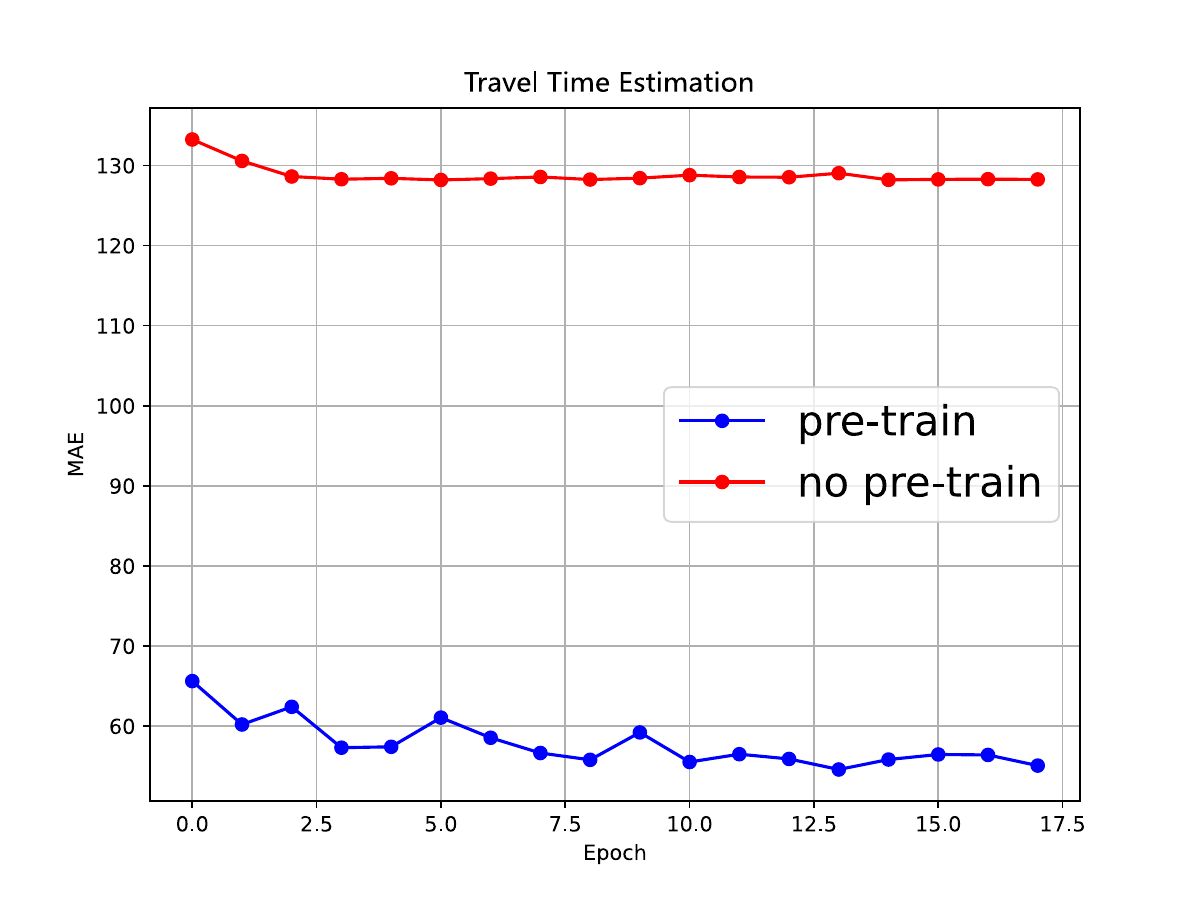}
            \label{fig: model_analysis_left}
        }
        \subfigure[RMSE]{
            \includegraphics[width=0.47\linewidth]{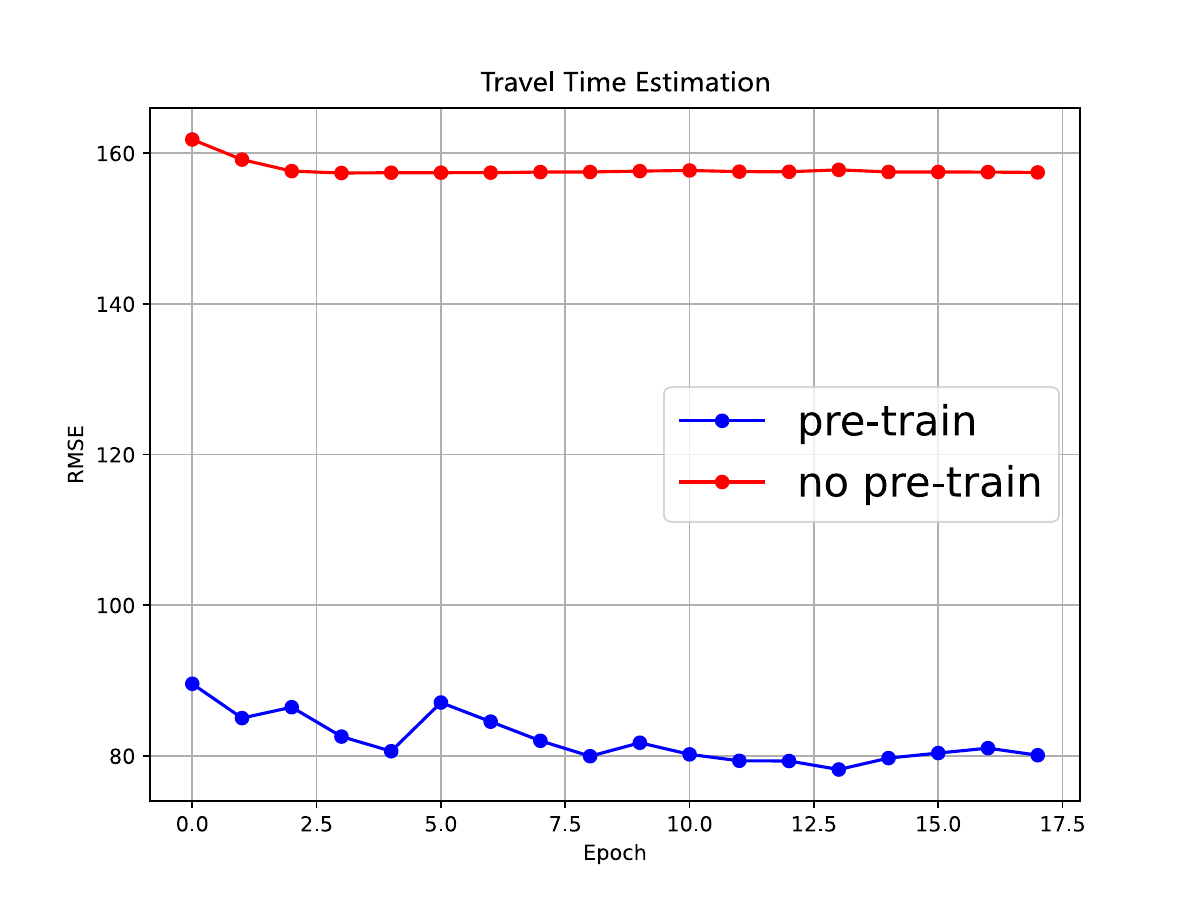}
            \label{fig: model_analysis_right}
        }
        \caption{Effect of Pre-training in Travel Time Estimation. }
        \label{fig: model_analysis}
    \end{figure}

    The results of these two strategies are presented in Figure~\ref{fig: model_analysis}. We observe that the model with pre-training consistently outperforms the end-to-end variant trained from scratch on the downstream task. The superior performance of the pre-trained model suggests that pre-training is a more effective strategy, as it enables the model to acquire comprehensive knowledge of trajectory patterns, resulting in faster convergence and reduced dependency on task-specific examples during fine-tuning. Furthermore, the results indicate that the pre-training stage captures generalizable trajectory features that are not tied to task-specific labels, contributing to enhanced performance across the downstream task.

\subsection{Ablation Study (RQ3)}\label{subsec:exp_rq3}

    \begin{table*}
        \caption{Ablation Study on Segment-Level and Trajectory-Level Tasks in Xi'an. }
        \label{tab:RQ3-Xi'an}
        \begin{tabular}{c|cc|cc|cc|cc}
            \toprule
             \multirow{2}*{Xi'an} & \multicolumn{2}{|c}{Road Speed Inference} & \multicolumn{2}{|c}{Road Label Classification} & \multicolumn{2}{|c}{Travel Time Estimation} & \multicolumn{2}{|c}{Destination Grid Prediction} \\
             & MAE $\downarrow$ & RMSE $\downarrow$ & Micro-F1 $\uparrow$ & Macro-F1 $\uparrow$ & MAE $\downarrow$ & RMSE $\downarrow$ & Acc@1 $\uparrow$ & Acc@5 $\uparrow$ \\
            \midrule
            MVTraj          & 2.5181 & 3.4623 & 0.8290 & 0.8159 & 54.9044 & 85.3847 & 0.6630 & 0.8154 \\ 
            w/o Inter-Modal & 2.5899 & 3.5219 & 0.8195 & 0.8160 & 86.9797  & 118.9468 & 0.5784 & 0.7594 \\ 
            w/o Grid View   & 3.1542 & 4.0976 & 0.7615 & 0.7468 & 73.1174 & 105.2485 & X & X \\
            w/o Align Loss  & 2.6353 & 3.5985 & 0.8247 & 0.8089 & 56.3331 & 86.8518 & 0.6446 & 0.7906 \\
            w/o MLM Loss    & 3.1959 & 4.1237 & 0.5200 & 0.4037 & 67.0827 & 99.3797 & 0.1552 & 0.3317 \\
            \bottomrule
        \end{tabular}
    \end{table*}

    We conduct an ablation study to assess the impact of various components of our method. The following model variants are examined: 
    \begin{itemize}
        \item w/o Inter-Modal: Omits inter-modal interaction attention.
        \item w/o Grid View: Excludes the grid view, utilizing only the other two views in JGRM.
        \item w/o Align Loss: Removes the alignment loss $\mathcal{L}_{align}^{multi}$, which aligns features across spatial views.  
        \item w/o MLM Loss: Excludes the masked language modeling loss $\mathcal{L}_{MLM}^{multi}$, which captures transition patterns in trajectories.
    \end{itemize}

    Experimental results on the Xi'an (Table~\ref{tab:RQ3-Xi'an}) and Chengdu datasets confirm that all components contribute positively to model performance. Due to space constraints, ablation results for Chengdu are provided in Appendix~\ref{subsec:appendix_ablation_chengdu}, with consistent findings across both segment- and trajectory-level tasks.

    Removing the inter-modal interaction attention (w/o Inter-Modal) significantly reduces performance, highlighting the importance of modeling interactions between spatial views, as the model struggles to integrate complementary information from GPS, route, and grid views.
    
    Excluding the grid view (w/o Grid View) severely impacts tasks requiring fine-grained spatial understanding, such as travel time estimation, by weakening the model’s ability to capture detailed spatial features in dense urban environments.
    
    Omitting the alignment loss (w/o Align Loss) leads to a moderate performance drop, indicating that while the model compensates to some extent, feature alignment still enhances prediction accuracy, especially in trajectory-level tasks.
    
    The largest performance degradation occurs when the MLM loss is removed (w/o MLM Loss), as this loss is critical for capturing contextual dependencies, improving generalization, and handling noisy or incomplete trajectory data.

\subsection{Model Efficiency (RQ4)}\label{subsec:exp_rq4}

    \begin{table}
        \caption{Efficiency Comparison on Xi'an.}
        \label{tab:RQ4-Xi'an}
        \begin{tabular}{cccc}
            \toprule
             & Model Size & Train Time & Inference Time \\
             & (MBytes) & (min/epoch) & (milliseconds) \\
            \midrule
            Random   & - & - & 0.374 \\
            Word2Vec    & 8.0 & 0.2 & 0.404 \\
            Node2Vec    & 7.6 & 0.2 & 0.328 \\
            Transformer & 14  & 1.5 & 0.940 \\
            BERT        & 433 & 3.5 & 1.260 \\
            START       & 94  & 17  & 4.700 \\
            JGRM        & 33  & 15  & 4.220 \\
            \midrule
            MVTraj      & 207 & 35 & 10.560 \\ 
            \bottomrule
        \end{tabular}
    \end{table}

    We further test the efficiency of MVTraj with several baseline models in terms of model size, training time, and inference time, as shown in Table~\ref{tab:RQ4-Xi'an}. While MVTraj exhibits a slightly longer inference time compared to the baseline methods, this is primarily due to the components of integrating multiple spatial views and performing real-time cross-modal interactions. However, it is important to highlight that the inference times for all models remain within the same order of magnitude (0-30 milliseconds). As a result, this minor difference in inference time has a negligible impact on the overall real-time performance and is unlikely to significantly affect the user experience during deployment.
    In terms of computational cost, MVTraj is more resource-intensive, with a larger model size and longer training time compared to the baseline models. However, these additional costs are justified by MVTraj’s substantial improvements in predictive accuracy. For applications where high precision is critical, this trade-off between efficiency and performance is well worth the investment. 

    In summary, although MVTraj incurs a slightly higher computational overhead, the efficiency trade-off is minimal when considering its significant performance gains, making it highly valuable for accuracy-driven real-world applications.

\section{Conclusion}

    In this paper, we introduced MVTraj, a novel self-supervised model for multi-view trajectory representation learning. MVTraj addresses two key challenges in this domain, multi-view alignment and knowledge fusion, by using GPS trajectories as a central bridge to integrate three complementary spatial views: GPS coordinates, routes, and POIs. This unified framework enables a more comprehensive understanding of movement patterns and spatial relationships, enhancing the flexibility and generalizability of the learned trajectory representations across diverse downstream tasks. 
    Extensive experiments on real-world datasets demonstrate that MVTraj significantly outperforms state-of-the-art methods, highlighting its effectiveness in extracting complementary knowledge from multiple spatial views. 

% \begin{acks}
% \end{acks}

\bibliographystyle{ACM-Reference-Format}
\bibliography{main_ref}

\appendix

\section{Appendices}

\subsection{Key Notions}

    To facilitate easy reference, Table~\ref{tab:notations} summarizes the key notations used throughout the paper.
    
    \begin{table}[h]
        \caption{Notations}
        \label{tab:notations}
        \centering
        \begin{tabular}{ll}
             \toprule
             Notation & Description \\
             \midrule
             $\mathcal{T}$ & Trajectory \\
             $pos_i$ & Position of the $i$-th point in a trajectory \\
             $t_i$ & Timestamp associated with the $i$-th point \\
             $G$ & Road network graph \\
             $\mathcal{V}$ & Set of vertices (road segments) in $G$ \\
             $\mathcal{E}$ & Set of edges (intersections) in $G$ \\
             $\mathbf{A}$ & Adjacency matrix of $G$ \\
             $\mathcal{T}^p$ & GPS trajectory \\
             $\mathcal{T}^r$ & Route trajectory \\
             $\mathcal{T}^g$ & Grid trajectory \\
             \bottomrule
        \end{tabular}
    \end{table}

\subsection{Experimental Setting}

\subsubsection{Details of Datasets}\label{subsec:appendix_datasets}

    The datasets used in this study were collected over a period of 15 consecutive days. For both datasets, the first 13 days are used for training, the 14th day for validation, and the 15th day for testing. The statistical details of these datasets are summarized in Table~\ref{tab:datasets}. 

    Each dataset consists of 13 distinct categories of points-of-interest (POI), representing a diverse range of urban functions. These categories include Dining, Scenery, Public Facilities, Shopping, Transportation, Education, Finance, Residential, Life Services, Sports, Healthcare, Government Offices, and Accommodation Services. This broad categorization ensures comprehensive coverage of different types of POI commonly found in urban environments. 

    \begin{table}[h]
        \caption{Datails of the Datasets}
        \label{tab:datasets}
        \centering
        \begin{tabular}{l|cc}
             \toprule
             Datasets & Chengdu & Xi'an \\
             \midrule
             $\#$ Trajectories &  100000 & 100000 \\
             Avg. GPS Trajectory Length ($m$) & 2829.16 & 2797.26 \\
             Avg. Route Trajectory Length & 15.26 & 15.96 \\
             Avg. Grid Trajectory Length & 18.36 & 37.08 \\
             Avg. Road Travel Speed ($m/s$ ) & 6.91 & 6.22 \\
             Avg. Trajectory Travel Time ($s$) & 426.31 & 467.47 \\
             \bottomrule
        \end{tabular}
    \end{table}

    \begin{table*}[h]
        \caption{Ablation Study on Segment-Level and Trajectory-Level Tasks in Chengdu. }
        \label{tab:RQ3-Chengdu}
        \begin{tabular}{c|cc|cc|cc|cc}
            \toprule
             \multirow{2}*{Chengdu} & \multicolumn{2}{|c}{Road Speed Inference} & \multicolumn{2}{|c}{Road Label Classification} & \multicolumn{2}{|c}{Travel Time Estimation} & \multicolumn{2}{|c}{Destination Grid Prediction} \\
             & MAE $\downarrow$ & RMSE $\downarrow$ & Micro-F1 $\uparrow$ & Macro-F1 $\uparrow$ & MAE $\downarrow$ & RMSE $\downarrow$ & Acc@1 $\uparrow$ & Acc@5 $\uparrow$ \\
            \midrule
            MVTraj          & 2.6972 & 3.7140 & 0.7206 & 0.7326 & 48.5581 & 71.8248 & 0.7927 & 0.9105 \\ 
            w/o Inter-Modal & 3.4545 & 4.5052 & 0.5053 & 0.4538 & 69.9533 & 96.8922 & 0.5472 & 0.7043 \\ 
            w/o Grid View   & 2.9213 & 3.9500 & 0.7041 & 0.7033 & 67.2398 & 94.1470 & X & X \\
            w/o Align Loss  & 2.8042 & 3.7839 & 0.7153 & 0.7311 & 49.9870 & 78.5322 & 0.7629 & 0.8973 \\
            w/o MLM Loss    & 3.4928 & 4.5430 & 0.4688 & 0.4053 & 77.1607 & 104.6159 & 0.1666 & 0.3504 \\
            \bottomrule
        \end{tabular}
    \end{table*}

\subsubsection{Details of Evaluation Metrics}\label{subsec:appendix_evaluation_metrics}

    We employ a variety of evaluation metrics to compare the performance of various methods: 
    
    \begin{itemize}
        \item Micro-F1: This metric aggregates the contributions of all classes to compute a single overall value. It sums the true positives (TP), false positives (FP), and false negatives (FN) across all classes to calculate overall precision and recall, and then combines them to compute the F1 score:
        \begin{displaymath}
            Micro-F1 = \frac{2 \times {Precision}_{all} \times {Recall}_{all}}{{Precision}_{all} + {Recall}_{all}}
        \end{displaymath}

        \item Macro-F1: This metric treats each class equally by calculating the F1 score for each class independently and then averaging these scores. It is given by: 
        \begin{displaymath}
            \begin{aligned}
                Macro-F1 = \frac{1}{N} \sum_{i=1}^{N} F1_{i}
            \end{aligned}
        \end{displaymath}
        where $F1_i$ is the F1 score for the $i$-th class, and $N$ represents the total number of classes. This metric is useful when class imbalance is a concern, as it gives equal weight to all classes, regardless of their size. 

        \item Mean Absolute Error (MAE): MAE measures the average magnitude of the errors between the predicted and true values, without considering their direction:
        \begin{displaymath}
            MAE = \frac{1}{n} \sum_{i=1}^{n} \vert y_i - \hat{y}_i \vert
        \end{displaymath}
        where $y_i$ is the true value, $\hat{y}_i$ is the predicted value, and $n$ is the total number of observations. MAE gives a direct interpretation of the average error.

        \item Root Mean Squared Error (RMSE): This metric represents the square root of the average squared differences between the predicted and true values. It emphasizes larger errors more than MAE due to the squared term: 
        \begin{displaymath}
            RMSE = \sqrt{\frac{1}{n} \sum_{i=1}^{n} ( y_i - \hat{y}_i )^2}
        \end{displaymath}

        \item Accuracy@k ($Acc@k$): This metric evaluates whether at least one correct prediction exists among the top $k$ predicted items for each instance:
        \begin{displaymath}
            Acc@k = \frac{1}{n} \sum_{i=1}^{n} \mathbb{I}(y_i \in \hat{y}_i^{(k)})
        \end{displaymath}
        where $y_i$ is the true label, $\hat{y}_i^{(k)}$ represents the set of top-$k$ predicted labels, and $\mathbb{I}(\cdot)$ is an indicator function, which returns 1 if the true label $y_i$ is within the top-$k$ predictions, and 0 otherwise. This metric is particularly useful in ranking tasks or recommendation systems.
        
    \end{itemize}

\subsubsection{Implementation Details}\label{subsec:appendix_implementation_details}

    All models, including MVTraj and the baseline models, are trained using the AdamW optimizer. The training is conducted for 70 epochs with a batch size of 64. 

    The loss function incorporates weights \(w_1\) and \(w_2\) as defined in Equation~\ref{eq:total_loss}, with values consistently set to 2 and 1, respectively, throughout all experiments. 

    To ensure fair and consistent comparisons with prior work~\cite{JGRM24,LightPath23,MMTEC2023,JCRLNT22}, we follow the same data preprocessing and splitting procedures. Specifically, we filter out road segments that are not covered by any GPS trajectories. Additionally, we remove trajectories that meet any of the following criteria: containing fewer than 10 or more than 100 road segments, fewer than 10 or more than 100 grid cells, or fewer than 10 or more than 256 GPS points.

    To further improve model robustness, we implement a masking mechanism during training. This mechanism uses a mask length of 2 and a masking probability of 20\%, which helps the model generalize better across diverse trajectory patterns.

    Finally, all hyperparameters and configurations are kept consistent across all experiments to ensure comparability and reproducibility of results.

\subsection{Additional Experiments}

\subsubsection{Ablation Study in Chengdu}\label{subsec:appendix_ablation_chengdu}

   Table~\ref{tab:RQ3-Chengdu} presents the results of the ablation experiments conducted in Chengdu. The findings mirror those in Xi'an, showing that MVTraj consistently achieves superior performance compared to its variants. 

\end{document}